\documentclass[pmlr]{jmlr}%
\usepackage{multirow}
\RequirePackage{graphicx}
 \usepackage{booktabs}

\usepackage{caption}
\usepackage{longtable}%

\makeatletter
\def\set@curr@file#1{\def\@curr@file{#1}} %
\makeatother
\usepackage[load-configurations=version-1]{siunitx} %

\theorembodyfont{\upshape}
\theoremheaderfont{\scshape}
\theorempostheader{:}
\theoremsep{\newline}

\jmlrvolume{298}
\jmlryear{2025}
\jmlrworkshop{Machine Learning for Healthcare}

\newcommand*\samethanks[1][\value{footnote}]{\footnotemark[#1]}

\title[TrajSurv]{TrajSurv: Learning Continuous Latent Trajectories from Electronic Health Records for Trustworthy Survival Prediction}

\author{\Name{Sihang Zeng}
        \Email{zengsh@uw.edu}\\ 
        \addr Department of Biomedical Informatics and Medical Education\\
        University of Washington\\
        Seattle, WA, USA
        \AND
        \Name{Lucas Jing Liu}
        \Email{jliu6@fredhutch.org}\\ 
        \addr Fred Hutch Cancer Center\\
        Seattle, WA, USA
        \AND
        \Name{Jun Wen}
        \Email{jun\_wen@hms.harvard.edu}\\ 
        \addr Department of Biomedical Informatics\\
        Harvard University\\
        Boston, MA, USA
        \AND
        \Name{Meliha Yetisgen}
        \Email{melihay@uw.edu}\\ 
        \addr Department of Biomedical Informatics and Medical Education\\
        University of Washington\\
        Seattle, WA, USA
        \AND
        \Name{Ruth Etzioni}\thanks{Co-senior authors.}
        \Email{retzioni@fredhutch.org}\\ 
        \addr Fred Hutch Cancer Center\\
        Seattle, WA, USA
        \AND
        \Name{Gang Luo}\samethanks
        \Email{luogang@uw.edu}\\ 
        \addr Department of Biomedical Informatics and Medical Education\\
        University of Washington\\
        Seattle, WA, USA}

\begin{document}

\maketitle

\begin{abstract}
Trustworthy survival prediction is essential for clinical decision making. Longitudinal electronic health records (EHRs) provide a uniquely powerful opportunity for the prediction.
However, it is challenging to accurately model the continuous clinical progression of patients underlying the irregularly sampled clinical features and to transparently link the progression to survival outcomes. To address these challenges, we develop TrajSurv, a model that learns continuous latent trajectories from longitudinal EHR data for trustworthy survival prediction. TrajSurv employs a neural controlled differential equation (NCDE) to extract continuous-time latent states from the irregularly sampled data, forming continuous latent trajectories. To ensure the latent trajectories reflect the clinical progression, TrajSurv aligns the latent state space with patient state space through a time-aware contrastive learning approach. To transparently link clinical progression to the survival outcome, TrajSurv uses latent trajectories in a two-step divide-and-conquer interpretation process. First, it explains how the changes in clinical features translate into the latent trajectory's evolution using a learned vector field. Second, it clusters these latent trajectories to identify key clinical progression patterns associated with different survival outcomes. Evaluations on two real-world medical datasets, MIMIC-III and eICU, show TrajSurv's competitive accuracy and superior transparency over existing deep learning methods.

\end{abstract}

\section{Introduction}

Accurate and transparent survival prediction is crucial for trustworthy clinical decision making \cite{Alabdallah_2025}. Longitudinal electronic health records (EHRs), which capture patients' evolving clinical status through irregularly sampled clinical features, provide rich temporal information for survival prediction \cite{lee2019dynamic}. 
Although deep learning models have been developed to leverage this information and enhance accuracy, two key challenges remain unresolved: accurately modeling continuous clinical progression and transparently linking that progression to survival outcomes \cite{xie2022deep}.

To accurately model the continuous clinical progression, it is important to aggregate irregularly sampled clinical features within a continuous-time framework in a clinically aligned way. Recurrent neural networks (RNNs) \cite{nagpal2021deep,lee2019dynamic} aggregate features at discrete times, potentially discarding the underlying continuous-time patterns. A recent approach, SurvLatentODE \cite{moon2022survlatent} leverages neural ordinary differential equations (NODEs) to aggregate features into continuous-time latent states. 
However, without high-quality supervision signals to guide the latent states at earlier time points, it may learn latent states' trajectories that do not consistently align with the patient's actual clinical progression, even if the ultimate latent state is shown to be clinically relevant.
This may lead to suboptimal modeling of clinical progression and suboptimal prediction accuracy.

To transparently link the clinical progression to the outcome, it is essential to provide an end-to-end interpretation of how changes in clinical features, i.e., feature velocities, lead to the outcome. However, existing models mainly interpret the contribution of clinical features' absolute values at observed time points \cite{lee2019dynamic}, but they do not fully explain how feature velocities relate to survival. This is a critical gap, as clinicians recognize the predictive value of feature velocities in longitudinal data besides their absolute values. For example, creatinine kinetics are used to define acute kidney injury \cite{Waikar2009CreatinineKA}.

To address these challenges, we introduce TrajSurv, which learns continuous latent trajectories from longitudinal EHR for trustworthy survival prediction. 
To ensure the accurate modeling of continuous clinical progression, TrajSurv uses a neural controlled differential equation (NCDE) \cite{kidger2020neural} to aggregate clinical feature changes over time into continuous latent trajectories, and designs a time-aware contrastive learning (TACL) objective to explicitly align the latent trajectories with actual clinical progression. 
TACL not only improves the prediction accuracy, but also learns clinically aligned latent trajectories that split the model into feature-to-trajectory and trajectory-to-outcome processes. This enables a divide-and-conquer approach for end-to-end transparency, leveraging a learned vector field and latent trajectory clustering. The implementation of TrajSurv is available at \url{https://github.com/zengsihang/TrajSurv}.

\subsection*{Generalizable Insights about Machine Learning in the Context of Healthcare}
\begin{itemize}
\setlength\itemsep{0em}
    \item \textbf{Accurate modeling of continuous clinical progression improves trustworthiness for longitudinal EHR analysis.} With NCDE and TACL, TrajSurv accurately models the continuous clinical progression, improving prediction accuracy and enabling transparency.
    \item \textbf{Divide-and-conquer approach improves end-to-end model transparency.} TrajSurv achieves its end-to-end transparency by splitting the model into feature-to-trajectory and trajectory-to-outcome steps for interpretation.
\end{itemize}

\begin{figure}[h]
    \centering
    \includegraphics[width=\linewidth]{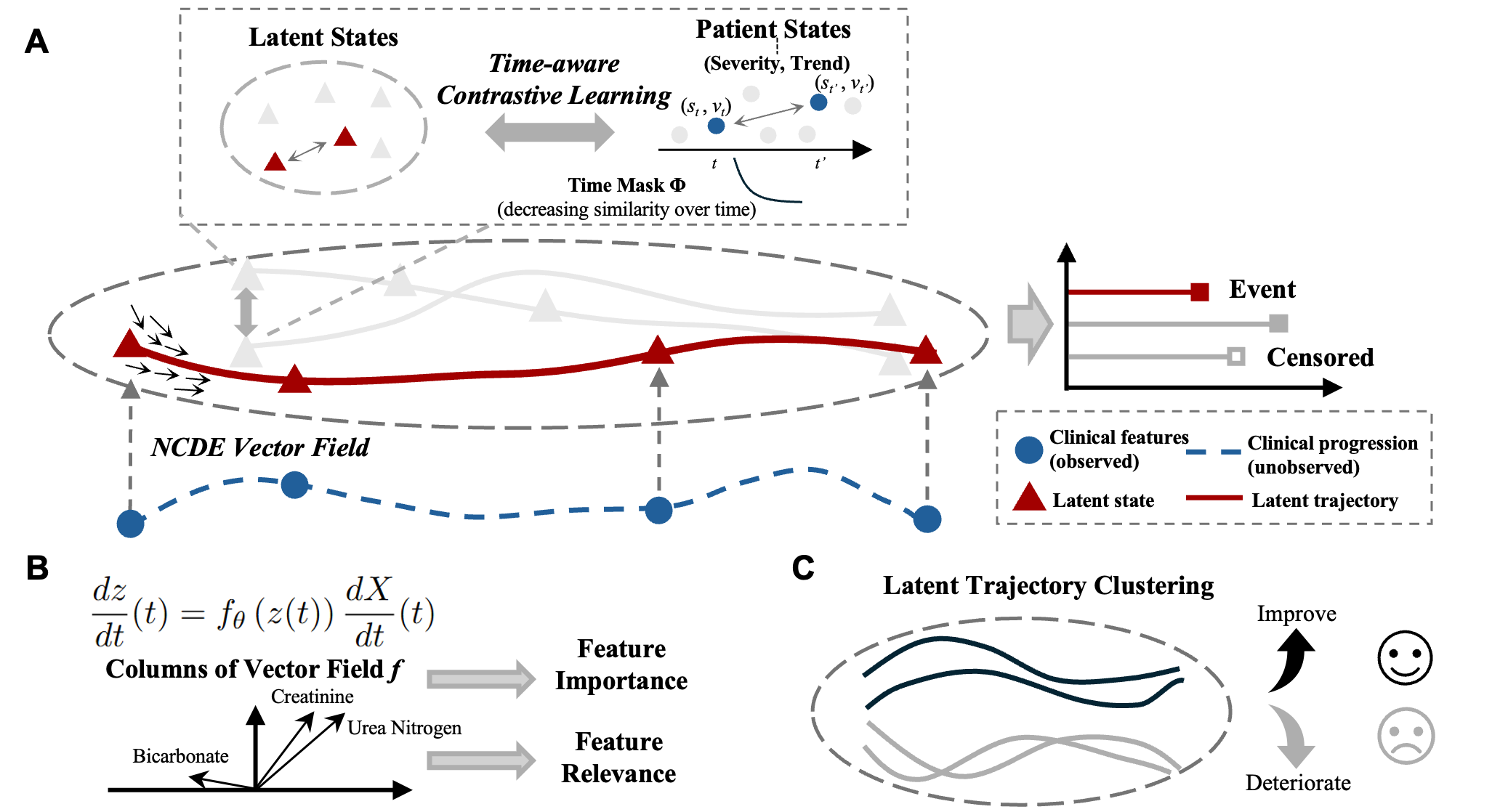}
    \caption{An illustration of TrajSurv and its two-step interpretation. (A) Model architecture. (B) TrajSurv's vector field interpretation. (C) TrajSurv's latent trajectory clustering.}
    \label{fig:main_fig}
\end{figure}

\section{Related Work}
Deep learning has demonstrated efficacy in survival prediction from longitudinal EHRs, capturing complex temporal dependencies. 
Recurrent Deep Survival Machines (RDSM) \cite{nagpal2021deep} and Dynamic-DeepHit \cite{lee2019dynamic} leveraged RNNs and aggregated irregularly sampled clinical features at discrete times, which may not accurately model the continuous clinical progression. 
Although Dynamic-DeepHit offered interpretability through feature importance, attention weights, and predicted risks, neither method fully explained how the clinical features' changes over time relate to survival outcomes. 
Recent continuous-time models, including NODEs \cite{chen2018neural}, ODE-RNN \cite{rubanova2019latent}, and NCDEs \cite{kidger2020neural}, map irregularly sampled time series into the continuously evolved latent states in the latent space through differential equations. 
SurvLatentODE \cite{moon2022survlatent} leveraged ODE-RNN to model time-varying clinical features. Still, the interpretation was limited to the last latent state, potentially lacking clinical relevance at earlier time points due to limited supervision, which may affect the model's accuracy and transparency.
CoxSig \cite{bleistein2024dynamical}, employing NCDEs, focused on theoretical advancements and offered limited insight into NCDE's latent trajectory.
In this study, we demonstrate how TrajSurv's continuous latent trajectories ensure accurate and transparent survival prediction.

\section{Methods}
\subsection{Problem Formulation}
\subsubsection{Survival Prediction}
We consider a right-censored survival dataset $\mathcal{D} = \{(\mathcal{X}^i, T^i, \delta^i)\}_{i=1}^N$, consisting of $N$ patients with longitudinal EHR data $\mathcal{X}^i$, time-to-event $T^i$, and event indicator $\delta^i$. For each patient $i$, the longitudinal EHR data $\mathcal{X}^i = \{(t^i_j, x^i_j)\}_{j=1}^{n_i}$ represents an irregularly sampled time series of clinical features, where $x^i_j \in \mathbb{R}^d$ is a d-dimensional clinical feature vector at time $t^i_j$, and $t^i_0=0$ is defined the same for all patients as a starting time point (e.g., time of admission). $\mathcal{X}^i$ is irregularly sampled, which means that the total number of observations $n_i$ and intervals between observations vary across patients, with clinical features containing missing values.
The time-to-event $T^i$ is defined by the time from the last observation $t^i_{n_i}$ to the event of interest (e.g., in-hospital death) or censoring (e.g., discharge). The event indicator $\delta^i = 1$ if the event occurs and $\delta^i = 0$ if censored.

The task of survival prediction is to model the survival distribution given $\mathcal{X}^i$ up to the last observation, predicting the hazard function:

\begin{equation}
\lambda(t | \mathcal{X}^i) = \lim_{\Delta t \to 0} \frac{P(t \leq T^i \leq t + \Delta t | T^i \geq t, \mathcal{X}^i)}{\Delta t} 
\end{equation}
or equivalently, the survival function $S(t | \mathcal{X}^i) = P(T^i > t | \mathcal{X}^i)$.

In this study, we aggregate longitudinal EHR data $\mathcal{X}^i$ in a shared latent space $\mathcal{Z}$, where the patient status at time $t$ is encoded as a latent state $z^i_t \in \mathcal{Z}$. The latent state $z^i_t$ summarizes the patient's historical information up to $t$ without knowing the clinical features after $t$, and the continuous latent trajectory $\tau^i=\{z^i_t, t\in[0, t^i_{n_i}]\}$ summarizes how latent states evolve in $\mathcal{Z}$. We seek to train a clinically aligned latent space $\mathcal{Z}$ for accurate and transparent survival prediction, where the continuous latent trajectory $\tau^i$ represents patients' clinical progression. For simplicity, we omit the superscript $i$ in the following sections, except where differentiation between patients is required. 

\subsubsection{Clinical Alignment}
To make the continuous latent trajectory clinically meaningful, we have to align the entire latent space with the clinical meaning of patient states, where a patient state captures the patient's current condition and the trend in the current condition. We define clinical alignment for latent states $z_t$ at a given time point $t$ as the property whereby close latent states correspond to similar patient conditions and their ongoing trends, while distant latent states correspond to distinct conditions or trends. This definition allows us to assess the local clinical relevance of the latent space at any specific time. However, directly comparing latent states at different time points does not provide a reliable measure of clinical alignment across these points. This is because the distance between $z_t$ and $z_{t'}$ (where $t \neq t'$) could arise from genuine differences in patient states or simply from drift over time. Therefore, we consider entire latent trajectories $\tau$, which inherently encode both time and the evolution of patient states. We define clinical alignment for latent trajectories as the property whereby similar trajectories represent similar evolutions of patient conditions, while dissimilar trajectories represent distinct evolutions. This definition focuses on the overall pattern of change over time, rather than point-to-point comparisons of individual latent states. As the trend information is inherently included in the evolution of patient conditions, it is not used as a separate metric to define clinical alignment for latent trajectories.

To practically capture clinical alignment, we utilize existing clinical assessments of severity as proxies for the patient’s condition. These assessments represent established clinical knowledge about patient conditions. Examples of such assessments include the Sequential Organ Failure Assessment (SOFA) score and the Acute Physiology And Chronic Health Evaluation (APACHE) score for intensive care unit (ICU) settings, and the Model for End-Stage Liver Disease (MELD) score for chronic liver disease. We denote the severity at time $t$ as $s_t$ and define the ongoing trend of severity as $v_t=(s_{t+\Delta t}-s_{t-\Delta t})/2\Delta t$, representing the changing rate of severity at $t$. The choice of $\Delta t$ is tailored to the specific clinical scenario, ensuring that the severity trend captures meaningful changes in the patient's condition over time, rather than merely reflecting noise from small time intervals. Both severity and its trend can be high-dimensional if the assessment involves multiple components (e.g., different components in the SOFA score).

\subsection{TrajSurv: Model Architecture}
In this section, we introduce the architecture of TrajSurv. TrajSurv uses an NCDE to map irregularly sampled clinical features in longitudinal EHR into continuous latent trajectories, where the last latent states are linked to the survival outcomes. We further design a time-aware contrastive learning objective (TACL) to align latent trajectories with actual clinical progression. Figure \ref{fig:main_fig}A shows an illustration of TrajSurv.

\subsubsection{Mapping Longitudinal EHR to Continuous Latent Trajectories}
TrajSurv uses NCDE as an encoder to aggregate temporal information from longitudinal EHR data. NCDE is a latent state-based method that learns continuous-time latent states driven by an irregularly sampled input process \cite{kidger2020neural}. Similar to previous work \cite{seedat2022continuous}, the use of NCDE in TrajSurv is motivated by its ability to capture the underlying continuous process from irregularly sampled clinical features, rather than discretely modeling the data (e.g., RNNs and transformers) or modeling through latent trajectories that only depend on initial values (e.g., NODEs). 

Specifically, TrajSurv maps longitudinal EHR $\mathcal{X}=\{(t_j, x_j)\}_{j=0}^n$ into a continuous latent trajectory $\tau=\{z_t, t\in [0,t_n]\}$. Following previous practices \cite{kidger2020neural,morrill2021neural}, the longitudinal EHR is interpolated into a continuous control signal $\{X_t, t\in [0, t_n]\}$ through the cubic Hermite splines with backward differences scheme such that $X_{t_j}=(x_j,t_j)\in \mathbb{R}^{d+1}$. 
The initial observation $X_0$ is mapped into a $d_z$-dimensional latent space using a feed forward network (FFN) $g_\phi:\mathbb{R}^{d+1}\rightarrow \mathbb{R}^{d_z}$. Subsequent continuous-time latent states are the solutions to an NCDE parameterized by a vector field $f_\theta:\mathbb{R}^{d_z}\rightarrow \mathbb{R}^{d_z\times(d+1)}$:
\begin{align}
\label{eq:ncde_int}
    z_t=z_{0}+\int^t_0 f_\theta(z_s)dX_s
\end{align}
where $z_0=g_\phi(X_0)$, $t\in[0, t_n]$, and the integral is a Riemann-Stieltjes integral. Therefore, the latent trajectory $\tau$ is the solution to an NCDE controlled by the underlying process of irregularly sampled clinical features. This transformation is realized through the vector field $f_\theta$, which maps the changes of clinical features $dX_t$ into the evolution of latent states. 

\subsubsection{Linking Latent Trajectory to Survival Outcome}
Because TrajSurv transforms irregularly sampled clinical features into evolved latent states, the last latent state $z_{t_n}$ accumulates the temporal information up to $t_n$. Therefore, $z_{t_n}$ is used as an aggregated predictor for the survival outcome. Similar to DeepSurv \cite{katzman2018deepsurv}, we use a nonlinear Cox proportional hazard model leveraging an FFN for more accurate prediction. In particular, the hazard function $\lambda(t|\mathcal{X})$ is decomposed into a baseline hazard $\lambda_0(t)$ and the exponential of a risk score $r$, where $r$ is derived from $z_{t_n}$ through an FFN $G_\eta: \mathbb{R}^{d_z}\rightarrow \mathbb{R}$:
\begin{equation}    \lambda(t|\mathcal{X})=\lambda(t|z_{t_n})=\lambda_0(t)\cdot r
    =\lambda_0(t)\cdot \exp\left({G_\eta(z_{t_n})}\right)
\end{equation}

We optimize TrajSurv's survival prediction using two loss functions. The first is the negative partial likelihood loss, commonly applied in Cox proportional hazard models \cite{cox1972regression}. For any $t \geq 0$, the risk set $R(t)$ is defined as the index set of patients still at risk (i.e., surviving equal or longer than $t$): $R(t) = \{i|T_i \geq t\}$. The negative log partial likelihood loss $\mathcal{L}_{PL}$ is then formulated as:

\begin{equation}
\mathcal{L}_{PL} = -\frac{1}{N_{\delta=1}} \sum_{i=1}^N \mathbb{I}(\delta^i = 1) \log \frac{\exp(r^i)}{\sum_{k \in R(T^i)} \exp(r^k)} \label{eq:lpl}
\end{equation}
where $N_{\delta=1}$ is the number of patients who experienced the event and $\mathbb{I}(.)$ is the indicator function. Minimizing $\mathcal{L}_{PL}$ corresponds to maximizing the risk score of patients who had events, relative to those who survived longer. To further improve the concordance of predicted risk scores between patients, similar to previous work \cite{wang2021combining}, we introduce a smoothed pairwise ranking loss $\mathcal{L}_{PR}$ to encourage patients with shorter survival times to have higher risk scores through pairwise ranking, defined as:

\begin{equation}
\mathcal{L}_{PR} = \frac{1}{N_{\delta=1}} \sum_{i=1}^N \sum_{k \in R(T^i)} \mathbb{I}(\delta^i = 1) \sigma(r^k - r^i) \label{eq:lpr}
\end{equation}
where $\sigma(x) = 1/(1 + \exp(-x))$ is the sigmoid function, providing a smoothed and differentiable approximation of the indicator function. The final survival loss $\mathcal{L}_{SURV}$ is the sum of the two loss functions.
\begin{equation}
\mathcal{L}_{SURV}=\mathcal{L}_{PL}+\mathcal{L}_{PR}
\end{equation}

\subsubsection{Clinical Alignment of Latent Trajectory}
Previous studies have shown that linking the last latent state $z_{t_n}$ to the survival outcome could learn clinically meaningful last states, where closer states represent similar outcomes \cite{moon2022survlatent}. 
However, because there is no explicit supervision on prior latent states, TrajSurv's latent trajectory $\tau$ is not guaranteed to be clinically aligned, i.e., the proximity of latent trajectories may not correspond to similar clinical progression, though they ultimately arrive at a clinically meaningful distribution. This may affect trustworthiness because the clinical progression information is not accurately and transparently captured. To ensure TrajSurv's latent trajectories reflect clinical progression, we design a TACL objective. 

While aligning two continuous-time processes, the latent trajectory and clinical progression, is challenging, the TACL objective simplifies this alignment by only aligning anchor latent states with high-quality labels while accounting for the time intervals between them. It is inspired by the anchor-based methods that select representative anchor points to enhance computational efficiency and reduce noise \cite{yang2025enhanced}, facilitating robust alignment. Although the alignment is performed at discrete time points for a latent trajectory, we expect it to propagate across the timeline due to the time-aware design and the continuous-time nature of the latent space, allowing for consistent alignment throughout the entire trajectory. 

We use latent states at observation time points as anchor latent states. For simplicity, we denote the set of anchor latent states as $Z=\{z_k\} = \bigcup_{i=1}^{N} \{z_{t_{j}}^{i}\}_{j=0}^{n_{i}} $. To avoid confusion, in the remainder of this section, $|Z|$ will refer to the number of anchor latent states in a single batch, and the subscripts will refer to distinct anchor latent states and their properties.

Specifically, we align anchor latent states $z_i$ across all patients in a batch with their corresponding severity $s_i$ and severity trend $v_i$ through contrastive learning \cite{khosla2021supervisedcontrastivelearning}. The contrastive learning guides anchor latent states with similar labels, which are severity and trend in our scenario, to be closer in the latent space. Because both severity and its trend are continuous variables, we employ a contrastive learning for regression framework called Rank-N-Contrast \cite{zha2024rank}. This framework learns the latent distance based on the rank of label distance. If we directly adopt Rank-N-Contrast, we have the negative log-likelihood loss:

\begin{equation}
\mathcal{L}_{RNC} = -\frac{1}{|Z|^2} \sum_{i=1}^{|Z|} \sum_{j=1}^{|Z|} \log \frac{\exp(\text{sim}(z_i, z_j) / \kappa_1)}{\sum_{z_k\in \mathcal{S}_{i,j}} \exp(\text{sim}(z_i, z_k) / \kappa_1)}
\end{equation}
where $\mathcal{S}_{i,j} = \{z_k | k \neq i, |s_i - s_k| + \delta |v_i - v_k| > |s_i - s_j| + \delta |v_i - v_j|\}$ is the set of latent states with larger label distance with $z_i$ than the label distance between $z_i$ and $z_j$, $\delta$ is a hyperparameter that balances the severity and its trend, $\kappa_1$ is a hyperparameter that controls the sensitivity of contrast, and $\text{sim}(\cdot, \cdot)$ is the similarity between two latent states.

However, this direct adoption does not account for the time intervals between latent states, which may inadequately capture the time-dependent nature of clinical progression. Therefore, we introduce a time mask $\Phi(t_i, t_j; \kappa_2) = \exp(-\frac{|t_i - t_j|}{\kappa_2})$ applied to the log-likelihood term in $\mathcal{L}_{RNC}$, forming our TACL objective. The time mask penalizes the contrasts between latent states separated by larger time intervals. This allows latent states with a large time interval to separate even if they share similar patient states. The hyperparameter $\kappa_2$ controls the strength of this penalty and can be adjusted for different clinical scenarios to reflect the time sensitivity of clinical progression. Formally, the TACL objective $\mathcal{L}_{TACL}$ is defined as:

\begin{equation}
\mathcal{L}_{TACL} = -\frac{1}{|Z|^2} \sum_{i=1}^{|Z|} \sum_{j=1}^{|Z|} \Phi(t_i, t_j; \kappa_2) \log \frac{\exp(\text{sim}(z_i, z_j) / \kappa_1)}{\sum_{z_k\in \mathcal{S}_{i,j}} \exp(\text{sim}(z_i, z_k) / \kappa_1)}
\end{equation}

Overall, TrajSurv's loss function is:
\begin{equation}
    \mathcal{L}=\mathcal{L}_{SURV}+\alpha \mathcal{L}_{TACL}
\end{equation}
where $\alpha$ adjusts the relative strength of clinical alignment to survival prediction.

\subsection{TrajSurv: Model Interpretation}
While the relationship between changes in clinical features and survival outcomes is complex,
TrajSurv's latent trajectory divides and conquers this process into two steps: mapping longitudinal EHR into continuous latent trajectories and linking latent trajectories to survival outcomes. The transparency can then be achieved by analyzing NCDE's vector field to understand how changes in clinical features lead to the evolution of latent trajectories, and by clustering latent trajectories to identify key clinical progression patterns linked to different survival outcomes. An illustration of model interpretation is shown in Figure \ref{fig:main_fig}B and C. 

\subsubsection{Step 1: Vector Field Interpretation}
\label{interpretation1}
The vector field $f_\theta$ maps the changes of clinical features into continuous latent trajectories. To further understand this transformation, we first transform the integral form of the NCDE \ref{eq:ncde_int} to the derivative form:
\begin{equation}
    \frac{dz}{dt}(t)=f_\theta\left(z(t)\right)\frac{dX}{dt}(t)
\end{equation}

For a specific time point $t$, the vector of latent velocity $\frac{dz}{dt}(t)$ is computed as the product of the matrix $f_{\theta}(z(t))$ and the vector of clinical feature velocity $\frac{dX}{dt}(t)$.
Expanding the matrix $f_\theta$ into column vectors, the latent velocity can be represented as the linear combination of the columns of $f_\theta$ weighed by the clinical feature velocity.
Therefore, each column of $f_{\theta}$ represents the influence of a unit change in a specific clinical feature on the magnitude and direction of the latent state's moving velocity. We analyze the columns of $f_{\theta}$ as follows:
\begin{itemize}
    \setlength\itemsep{0em}
    \item \textbf{Feature Importance:} The magnitude of each column indicates the importance of the corresponding clinical feature in driving latent state evolution. Features with larger column magnitudes have a greater impact on the evolution of the latent state.
    \item \textbf{Feature Relevance:} The cosine similarity between two columns captures the relevance of the corresponding features in driving latent states evolution. If the cosine similarity is high, it suggests that similar changes in these features tend to move the latent state in similar directions. Conversely, if the cosine similarity is low, it indicates that similar changes in these features tend to move the latent state in opposing directions.
\end{itemize}

This approach provides an elegant and interpretable framework for understanding how changes in clinical features drive latent state evolution and produce the latent trajectory, through the feature importance and relevance. 
Practically, instead of analyzing the vector field at specific times for specific patients, we estimate the average vector field $\bar{f}_{\theta}$ by sampling latent states $z(t)$ across patients and time points and averaging their vector field $f_\theta(z(t))$. This average field provides an overall perspective on feature importance and relevance. We note that these average patterns may not generalize to individual cases and should be interpreted cautiously.

\subsubsection{Step 2: Latent Trajectory Clustering}
\label{interpretation2}
Trained with the TACL objective, TrajSurv's continuous latent trajectory is expected to be clinically aligned. This means that similar latent trajectories, reflecting the evolution of underlying patient states, can correlate with similar clinical progression and survival outcomes. To investigate this, we employ dynamic time warping (DTW) \cite{muller2007dynamic}, a time series clustering method that finds optimal alignments between sequences of different lengths and outputs distances. Using DTW, we cluster latent trajectories ($\tau_i, i=1,..., N$) into $C$ distinct groups. We visualize the cluster centroids—derived using the DTW barycenter averaging algorithm—in the latent space. We compute the average severity trajectory over time for each cluster, aiming to reveal key clinical progression patterns. We further assess the association between the clusters and survival outcomes using Kaplan-Meier (KM) curves, hypothesizing that latent trajectory patterns may serve as prognostic indicators.

\section{Experiments}
In this section, we briefly introduce the experiment settings and refer readers to the Appendix for more details.
\subsection{Dataset and Setup}
We evaluated TrajSurv's survival prediction performance on two real-world EHR datasets in the ICU setting, MIMIC-III \cite{johnson2016mimic} and eICU \cite{pollard2018eicu}. Similar to prior work \cite{moon2022survlatent}, the prediction task is to estimate time to in-hospital mortality based on longitudinal EHR data in the first 36 hours of admission. This prediction task is crucial for timely care and optimal resource allocation in ICU \cite{moon2022survlatent}.

In ICU, the SOFA score assesses the performance of six organ systems (respiration, coagulation, liver, cardiovascular, neurologic, and renal), with each component ranging from 0 to 4 and higher scores indicating more severe organ dysfunction. The overall SOFA score is the sum of these 6 components and higher scores indicate a higher risk of ICU mortality \cite{shafigh2024prediction}. In the experiments, we used the SOFA score and its six components as $s_t$, representing patient conditions over time. SOFA scores $s_t$ were processed hourly from forward-imputed data, and SOFA trends $v_t$ were computed by the SOFA's average changing rate over 4-hour intervals ($\Delta t=2)$ to smooth fluctuations and capture meaningful trends. 

\subsection{MIMIC-III and eICU Cohort}
The MIMIC-III cohort included 20,258 hospital admissions and 53 clinical features in 1-hour resolution. The eICU cohort included 116,503 hospital admissions and 53 clinical features in 1-hour resolution. Both cohorts featured irregularly sampled time series data. To ensure computational efficiency, the time series data was pre-processed to only include hourly time points with more than 20 observed features (See Appendix \ref{app:generalization} and \ref{app:exp}). For both cohorts, $t_0$ was defined as the time of hospital admission, and the time-to-event $T$ was the time from the last record $(t_n, x_n); t_n\le 36$ to in-hospital mortality. Patients who survived were right censored at discharge. Clinical features were standardized. Each dataset was randomly split into training (70\%), validation (10\%), and test sets (20\%).

\subsection{Comparison Methods}
We compared the survival prediction performance of TrajSurv with three machine learning models, one clinical model, and four deep learning models. 
The machine learning models included the Cox proportional hazards model with 
elastic net penalty (CoxPH) \cite{cox1972regression}, gradient-boosted Cox 
proportional hazards model (Boosting) \cite{cox1972regression,friedman2001greedy}, 
and random survival forest (RSF) \cite{ishwaran2008random}. 
For the clinical model, we computed the longitudinal overall SOFA scores every 4 hour to create an aggregated 9-dimensional SOFA feature, and applied the Cox proportional hazards model (SOFA) for survival prediction.
In the deep learning category, we evaluated models designed 
for survival prediction using longitudinal EHR data, including RDSM \cite{nagpal2021deep}, SurvLatent ODE (SLODE) \cite{moon2022survlatent}, 
and Dynamic-DeepHit (DDH) \cite{lee2019dynamic}.

\subsection{Evaluation Metrics}
We evaluated survival prediction performance using three commonly used time-dependent metrics: the concordance index (C-index), the time-dependent Brier score, and the dynamic area under the curve (AUC). The C-index and dynamic AUC measure the model's discrimination ability to distinguish patient risks.
The Brier score assesses both the model's discrimination ability and its calibration. 
Specifically, we implemented the metrics using the \textit{concordance\_index\_ipcw}, \textit{brier\_score}, and \textit{cumulative\_dynamic\_auc} functions from scikit-survival module \cite{sksurv}.
We computed the average across quartiles of follow-up times in the dataset for these metrics. We reported the average performance across 5 runs with different random seeds. Hyperparameters were tuned on the validation set.

For TrajSurv, we further evaluated the clinical alignment between latent states and patient states by computing Spearman's correlation between latent distance and SOFA or SOFA trend distance at each time point.

\section{Results on Real Data}
\subsection{Survival Prediction Performance}
\begin{table}[h]
\centering
\caption{Survival Prediction Performance Comparison, mean $\pm$ std. \textbf{Bold} values indicate the best performance among all models or deep learning models for a given metric.}
\label{tab:performance}
\resizebox{\textwidth}{!}{ %
\begin{tabular}{l l ccc ccc}
\toprule
& & \multicolumn{3}{c}{MIMIC-III} & \multicolumn{3}{c}{eICU} \\
\cmidrule(lr){3-5} \cmidrule(lr){6-8}
& Model & C-index & Brier & AUC & C-index & Brier & AUC \\
\midrule
\multirow{3}{*}{ML} & CoxPH & 0.706 $\pm$ 0.012 & 0.058 $\pm$ 0.002 & 0.698 $\pm$ 0.012 & 0.751 $\pm$ 0.007 & 0.044 $\pm$ 0.001 & 0.733 $\pm$ 0.007 \\
& Boosting & 0.766 $\pm$ 0.014 & 0.054 $\pm$ 0.002 & 0.754 $\pm$ 0.014 & 0.781 $\pm$ 0.005 & 0.043 $\pm$ 0.001 & 0.761 $\pm$ 0.005 \\
& RSF & 0.784 $\pm$ 0.013 & \textbf{0.053 $\pm$ 0.002} & 0.774 $\pm$ 0.014 & 0.804 $\pm$ 0.004 & \textbf{0.042 $\pm$ 0.001} & 0.783 $\pm$ 0.004 \\
\midrule
Clinical & SOFA & 0.754 $\pm$ 0.012 & 0.053 $\pm$ 0.003 & 0.729 $\pm$ 0.013 & 0.755 $\pm$ 0.006 & 0.044 $\pm$ 0.001 & 0.726 $\pm$ 0.006 \\
\midrule
\multirow{7}{*}{DL} & RDSM & 0.772 $\pm$ 0.012 & 0.061 $\pm$ 0.002 & 0.757 $\pm$ 0.012 & 0.796 $\pm$ 0.004 & 0.048 $\pm$ 0.001 & 0.779 $\pm$ 0.004 \\
& SLODE & 0.762 $\pm$ 0.018 & 0.064 $\pm$ 0.004 & 0.745 $\pm$ 0.020 & 0.780 $\pm$ 0.007 & 0.045 $\pm$ 0.000 & 0.756 $\pm$ 0.007 \\
& DDH & 0.768 $\pm$ 0.014 & 0.072 $\pm$ 0.005 & 0.748 $\pm$ 0.014 & 0.818 $\pm$ 0.005 & 0.051 $\pm$ 0.002 & 0.796 $\pm$ 0.005 \\
& \textbf{TrajSurv} & \textbf{0.803 $\pm$ 0.011} & \textbf{0.056 $\pm$ 0.002} & \textbf{0.790 $\pm$ 0.013} & \textbf{0.823 $\pm$ 0.006} & \textbf{0.044 $\pm$ 0.001} & \textbf{0.803 $\pm$ 0.006} \\
\bottomrule
\end{tabular}
}
\end{table}

Table \ref{tab:performance} presents the survival prediction performance across comparison models and TrajSurv.
Compared to existing models, TrajSurv achieved a higher C-index and dynamic AUC and comparable Brier score, indicating improved discrimination and competitive calibration performance.  
(For more details, see Appendix)

\paragraph{Ablation Study} To examine how our additional objectives improve the performance over vanilla NCDE, we conducted ablation studies on MIMIC-III by removing the time mask in $\mathcal{L}_{TACL}$ (A1), removing the entire $\mathcal{L}_{TACL}$ (A2), and removing both $\mathcal{L}_{TACL}$ and $\mathcal{L}_{PR}$ (A3). 
As shown in Figure \ref{fig:ablation}A and B, comparing A2 and A3, we found that adding $\mathcal{L}_{PR}$ improves the performance over vanilla NCDE. Comparing A1 and A2, we found that contrastive learning slightly improves the discrimination ability, potentially due to additional supervision for the latent states. TrajSurv outperforms both A1 and A2, implying the validity of our model design involving the time-dependent nature of clinical progression. 

\paragraph{Cross-Cohort Generalization} To assess out-of-distribution generalization, we performed a cross-cohort evaluation by training TrajSurv on the full MIMIC-III training set and testing it on a random sample of 4,000 patients from the eICU dataset. The model achieved a C-index of $0.760$ and a Brier score of $0.038$, demonstrating robust performance on unseen data from different hospital systems.

\subsection{Clinical Alignment Performance}
\begin{figure}[h]
    \centering

    \includegraphics[width=\linewidth]{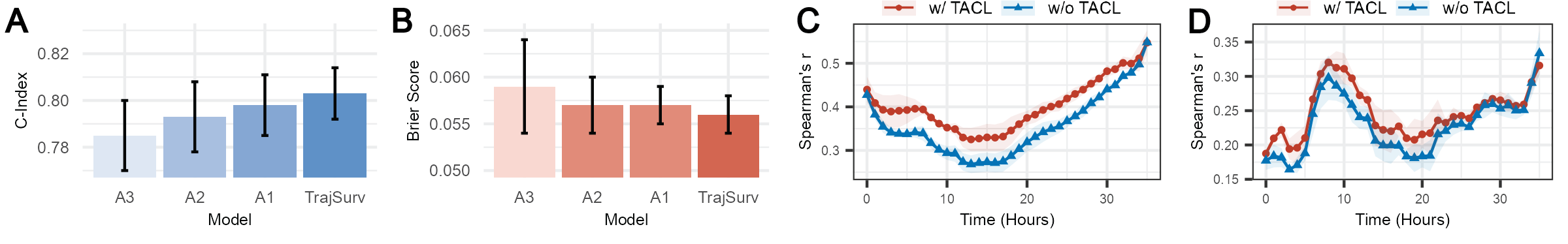}
    \caption{Ablation study showing improved (A) C-index and (B) Brier score. Spearman's correlation between latent distance and label distance, (C) SOFA distances, and (D) SOFA trend distances, with or without the TACL objective.}
    \label{fig:ablation}
\end{figure}

Figure \ref{fig:ablation}C and D shows how the correlations change over time with and without TACL on MIMIC-III data. Without TACL, the model has relatively high correlations with SOFA and SOFA trend at the end of the 36-hour period, while they drop significantly in the middle. This suggests that despite the clinical alignment of the last latent state, the evolution of latent states does not align with the clinical progression. Trained with the TACL objective, TrajSurv retains the correlations at the end, and has significantly higher correlations with SOFA and slightly higher correlations with SOFA trend in the middle of the 36 hours. This suggests that latent states across the time horizon are more clinically aligned with patient states. Therefore, the evolution of latent states better reflects the clinical progression.

\subsection{Interpretation}
In this section, we validate TrajSurv's two-step interpretation process with known clinical knowledge on the test set of MIMIC-III data at the population level.

\subsubsection{Feature Importance and Relevance from TrajSurv's Vector Field}
\begin{figure}[h]
    \centering
    \includegraphics[width=\linewidth]{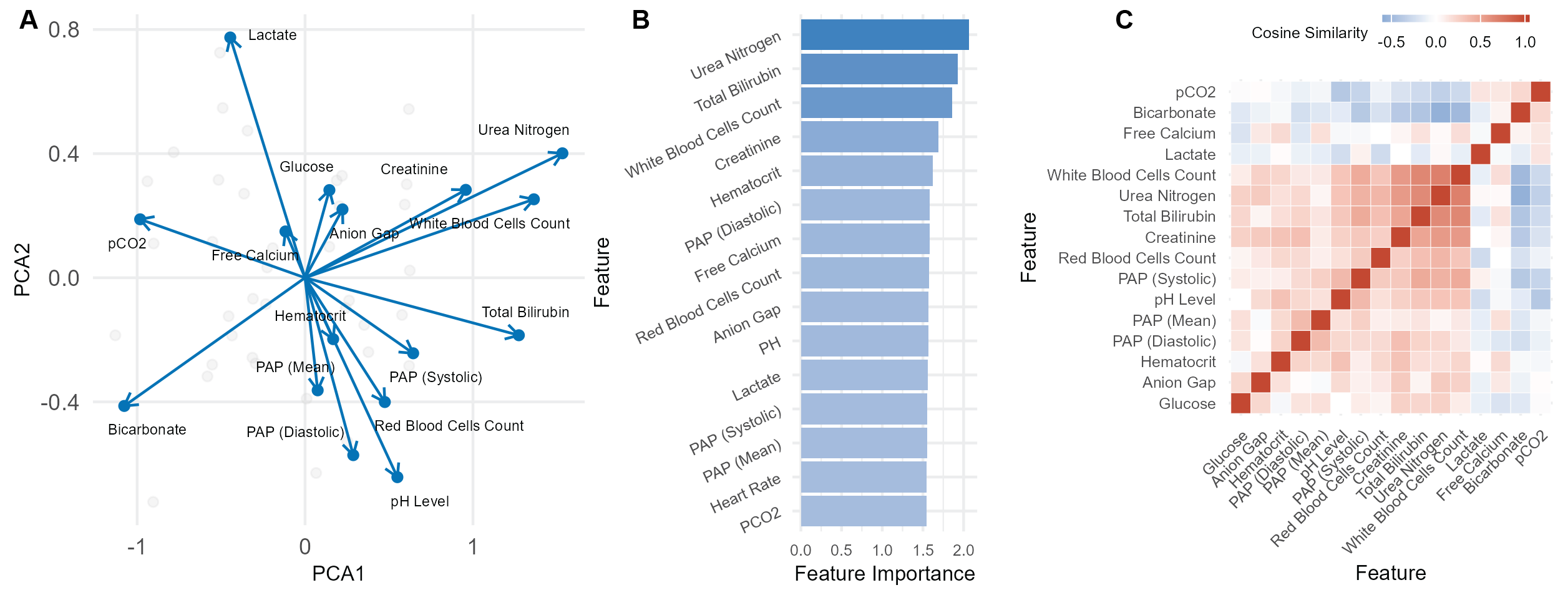}
    \caption{Feature importance and relevance derived from TrajSurv's vector field. (A) Columns of the average vector field; (B) Features with top 15 importance; (C) Cosine similarities between selected features.}
    \label{fig:vector_field}
\end{figure}

From the columns of $\bar{f}_\theta$ (Figure \ref{fig:vector_field}A), we obtained feature importance and feature relevance following the procedure in section \ref{interpretation1}. As shown in Figure \ref{fig:vector_field}B, the top 15 influential features are ranked by the magnitude of the columns in the average vector field. Among the most important are blood urea nitrogen, total bilirubin, white blood cell count, creatinine, and hematocrit. These findings are consistent with previous ICU studies identifying these as significant indicators of patient states \cite{chia2021explainable,iwase2022prediction}. For instance, blood urea nitrogen, creatinine, hematocrit, lactate, and PH were identified among the top 6 significant clinical features in a previous study \cite{chia2021explainable}. The feature importance derived from the vector field provides a dynamical perspective, suggesting that changes in these features drive latent state shifts more rapidly than others. 

Additionally, we plotted a heatmap of cosine similarities between columns of the vector field to reveal specific relationships among features (Figure \ref{fig:vector_field}C), providing an understanding of feature relevance in driving latent state transitions. For instance, creatinine and urea nitrogen display high positive cosine similarity, suggesting that their similar change patterns result in latent state shifts in nearly identical directions. This alignment is clinically consistent, as both are indicators of kidney function, where elevated levels may indicate renal impairment \cite{hosten1990bun}.
In contrast, urea nitrogen and bicarbonate show a high negative cosine similarity, meaning their parallel increases or decreases drive latent states in opposing directions. This finding aligns with evidence that urea nitrogen and bicarbonate are often inversely related in certain conditions \cite{papadoyannakis1984effect,balakrishnan2011blood}. For example, bicarbonate supplementation is associated with reduced blood urea nitrogen levels in chronic renal failure patients \cite{papadoyannakis1984effect}.

These relationships derived from TrajSurv's vector field confirm known clinical associations while illustrating how TrajSurv captures the dynamical interdependencies between features that influence latent state evolution. We found these interpretations to be robust across different random data splits. For instance, blood urea nitrogen, total bilirubin, and white blood cell count were consistently ranked as top-5 important features, and the cosine similarity between creatinine and urea nitrogen remained stable at $0.60\pm0.06$ across three different random seeds. Note that the feature importance and relevance were data-driven and should be interpreted with caution.

\begin{figure}[h]
    \centering
    \includegraphics[width=\linewidth]{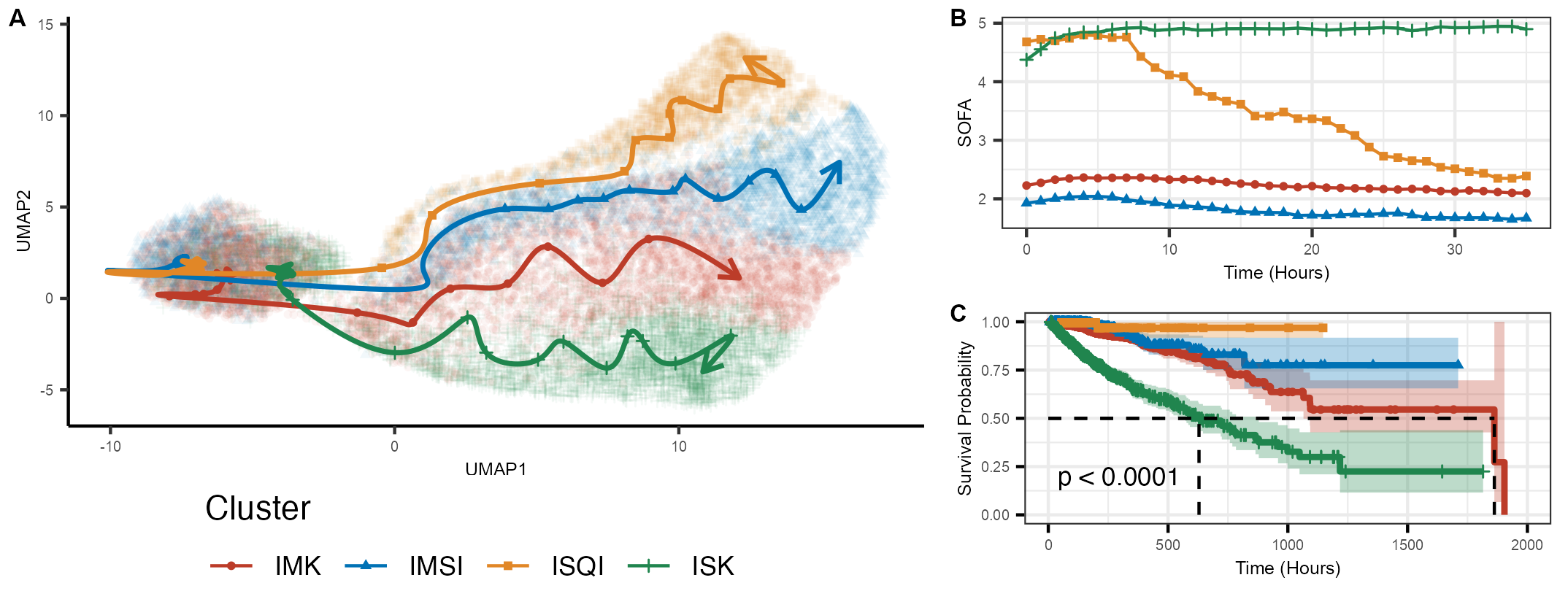}
    \caption{Clustering of latent trajectories with (A) the centroids of clusters; (B) the average SOFA trajectories of clusters; and (C) survival curves of clusters.}
    \label{fig:trajectory}
\end{figure}
\subsubsection{Latent Trajectory Clustering}
Clustering latent trajectories using DTW, we obtained four clusters. Figure \ref{fig:trajectory}A shows the centroids of four clusters in the latent space, dividing the space into subdomains. Examining the average SOFA score trajectories (Figure \ref{fig:trajectory}B) of these clusters reveals four distinct disease progression patterns. ISQI cluster starts with high SOFA scores that decrease rapidly (initially severe, quickly improving - ISQI). ISK cluster maintains a consistently high SOFA score with a slight increase early on, indicating an initially severe condition that gradually worsens (initially severe, keep - ISK). IMK cluster and IMSI cluster exhibit lower SOFA scores, with scores remaining stable over time for the former (initially mild, keep - IMK) and slowly improving for the latter (initially mild, slowly improving - IMSI). These clusters suggest that TrajSurv’s continuous latent trajectories represent different clinical progression patterns in the ICU.

The KM survival curve for each cluster reveals distinct survival patterns aligned with the clinical progression identified in the latent trajectories, hence stratifying risks. The ISQI subgroup, characterized by severe initial conditions that rapidly improve, shows the highest survival probability over time, suggesting that patients with early improvement after critical onset are less likely to experience adverse outcomes. In contrast, the ISK subgroup, with consistently high severity and a gradually worsening trend, displays the lowest survival probability, indicating a sustained high risk. The IMSI and IMK subgroups, representing milder initial conditions with slow improvement or stability, show intermediate survival outcomes. The statistically significant separation among the KM curves highlights the strength of TrajSurv’s continuous latent trajectories in capturing meaningful clinical progression patterns, effectively stratifying risk based on the evolved patient conditions.

\subsection{Case Study}
\begin{figure}[h]
    \centering
    \includegraphics[width=\linewidth]{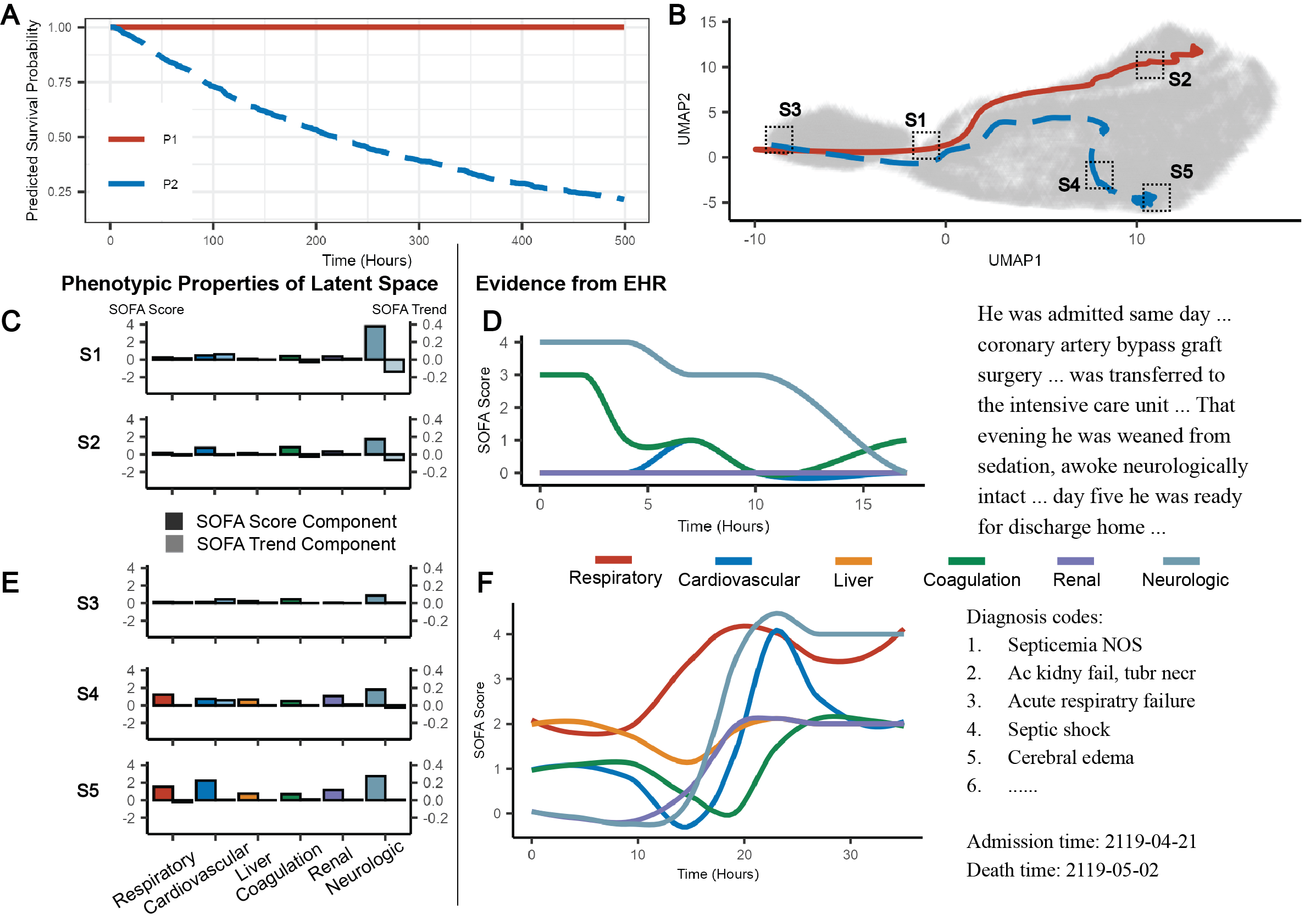}
    \caption{Case studies of two patients (P1 and P2). (A) Predicted survival probability after the last observation; (B) Continuous latent trajectories until the last observation; (C) Average phenotypic properties of S1 and S2 of the latent space; (D) Evidence from EHR for P1, including the ground truth SOFA trajectory and the discharge note; (E) Average phenotypic properties of S3, S4, and S5; (F) Evidence from EHR for P2, including the ground truth SOFA trajectory and the Diagnosis codes.}
    \label{fig:case_study}
\end{figure}
A case study of two patients in MIMIC-III demonstrates TrajSurv’s patient-level interpretation of how latent trajectories lead to predictions consistent with actual clinical progression. 

For patient P1, TrajSurv predicts a favorable survival outcome after the last observation (Figure \ref{fig:case_study}A). This prediction is supported by several aspects of P1’s latent trajectory. First, as shown in Figure \ref{fig:case_study}B, P1’s latent trajectory falls within the ISQI subgroup, indicating a generally positive prognosis. 
Additionally, P1's latent trajectory sequentially goes through regions S1 and S2 on the latent space, where the phenotypic properties of latent space (see Appendix for the calculation) show severe but improving neurologic symptoms at S1 and mild symptoms at S2 (Figure \ref{fig:case_study}C). This is consistent with P1's actual SOFA trajectory and discharge note, confirming neurologic improvement post-coronary artery bypass surgery on the first admission day (Figure \ref{fig:case_study}D).

For patient P2, TrajSurv predicts a poor survival outcome following the last observation (Figure \ref{fig:case_study}A). This result is supported by P2’s latent trajectory (Figure \ref{fig:case_study}B). 
P2’s trajectory shows a sudden shift from the IMSI subdomain to ISK subdomain, suggesting an abrupt deterioration mid-visit. 
P2's trajectory passes through S3, S4, and S5 regions, which indicate initial presentation with mild symptoms (S3) that progress to multiple organ dysfunction (S4 and S5).
In the absence of discharge notes, we compared this interpretation to P2’s actual SOFA trajectory, diagnosis codes, and actual survival time, showing consistency with a sepsis diagnosis and rapid decline leading to death within 300 hours.

This case study suggests TrajSurv’s potential to provide clinicians with interpretable insights consistent with patient outcomes and clinical progression.

\section{Discussion}

In this study, we propose TrajSurv, a method for survival prediction that learns continuous latent trajectories from longitudinal EHR. TrajSurv leverages NCDE to capture the continuous process of evolving clinical features, mapping this evolution to continuous latent trajectories. We further incorporate a TACL objective to ensure the clinical alignment of these trajectories, improving both accuracy and transparency. Paired with a two-step divide-and-conquer interpretation process, TrajSurv enables trustworthy survival prediction, achieving competitive prediction performance and superior transparency.

With increasing emphasis on the need for reliable and transparent AI systems in healthcare, TrajSurv provides a viable direction for the transparent modeling of longitudinal EHR by learning clinically aligned continuous latent trajectories and interpreting the model through a divide-and-conquer approach. The divide-and-conquer interpretation process provides a dynamical perspective on how the temporal changes of clinical features link to the outcome, enriching existing interpretation methods and fostering trust in the model's outputs. By providing a clear and visual representation of clinical progression, TrajSurv contributes to the development of trustworthy AI tools for clinical use.

A key innovation in TrajSurv lies in its TACL objective. This objective works as a soft regularization to the latent space by explicitly aligning the latent trajectory with actual clinical progression while respecting its time-dependent nature. By contrasting latent representations of patient states at different time points, TACL encourages the model to learn trajectories that are not only predictive of survival but also clinically meaningful. This enables clinically meaningful vector field analysis and cluster patterns in our two-step interpretation process. Through our ablation, we found that TACL improved both performance and clinical alignment. 

The patient's clinical progression is a complex and dynamic process, and TrajSurv’s TACL offers insights into how latent trajectories align with this process. Our analysis reveals that TACL learns latent states that correspond more closely to severity levels than to severity trends (Figure \ref{fig:ablation}C and D). This finding is consistent with prior research showing that mean and peak SOFA scores are stronger ICU prognosticators than changes in SOFA ($\Delta$-SOFA) \cite{ferreira2001serial}, suggesting that TACL may prioritize the most salient dimensions of multidimensional patient states. Future research may further refine this alignment by incorporating more comprehensive patient state definitions or advancing alignment techniques. As one of the early efforts, TrajSurv establishes a framework for transparent longitudinal EHR modeling by quantifying patient states, aligning latent trajectories with clinical progression, and interpreting with a divide-and-conquer approach.

Our work builds upon and extends previous research utilizing NCDEs. While studies like TE-CDE \cite{seedat2022continuous} and CoxSig \cite{bleistein2024dynamical} have employed NCDEs for tasks like continuous-time counterfactual prediction and dynamical survival prediction, they have largely underexplored the rich information contained within the latent trajectory and vector field. TrajSurv specifically leverages NCDE's capacity to represent the continuous nature of clinical progression transparently and quantitatively, linking clinical feature evolution to the latent trajectory. Crucially, the feature importance and relevance derived from the NCDE vector field provide a novel and elegant way to interpret how changes in different clinical features jointly influence changes in the latent space. For example, similar changes in creatinine and urea nitrogen lead to the similar transition of latent states.

Finally, TrajSurv holds considerable potential for clinical translation. Its accurate survival prediction and risk stratification can facilitate clinical resource allocation and decision-making. Critically, as shown in the case study, the interpretable outputs provided by TrajSurv can empower clinicians by providing insights into the prediction process, fostering trust and acceptance in clinical settings. While comprehensive clinical validation across diverse cohorts remains necessary, we believe TrajSurv represents a significant step toward trustworthy AI tools in clinical practice and can inspire the development of future explainable AI models for longitudinal EHR data.

\paragraph{Limitations}

Our model design relies on an existing clinical assessment like SOFA as the supervision signal. While these assessments offer a practical way to quantify patient states, they do not fully represent patient states and reduce the model flexibility in various clinical contexts, as not all clinical contexts provide such assessments. Future research may explore purely data-driven methods to ensure clinical alignment or define patient states more comprehensively using more adaptable labels, such as knowledge graphs or key clinical variables, rather than pre-existing assessments. Furthermore, future research is needed to understand the clinical alignment between latent trajectories and clinical progression from a more holistic view. Finally, population-level properties may not fit individual cases, and we leave individual-level vector field analysis for future studies.

\bibliography{main}

\newpage
\appendix
\section*{Appendix A. Additional Results}
\setcounter{figure}{0} %
\setcounter{table}{0}
\renewcommand{\thefigure}{S\arabic{figure}} %
\renewcommand{\thesubsection}{A.\arabic{subsection}}
\renewcommand{\thetable}{S\arabic{table}}
\renewcommand{\theHtable}{S\arabic{table}}%
\renewcommand{\theHfigure}{S\arabic{figure}}%

\subsection{Model Calibration}
\begin{figure}[h]
    \centering
    \includegraphics[width=\linewidth]{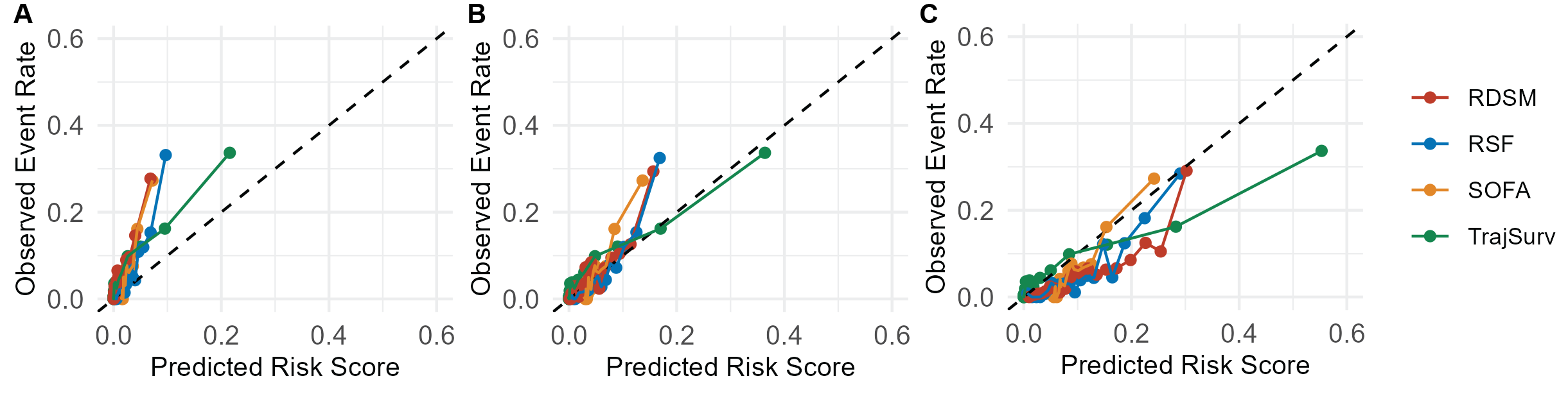}
    \caption{Calibration plot comparing TrajSurv with RSF, SOFA, and RDSM at (A) 25\% quartile, (B) median, and (C) 75\% quartile follow-up time.}
    \label{fig:Calibration}
\end{figure}

To assess TrajSurv's calibration, we generated calibration plots across quartiles of follow-up times on MIMIC-III data. Figure \ref{fig:Calibration} illustrates that TrajSurv’s predicted risk scores align well with observed outcomes at the median follow-up time but show slight underestimation at the 25\% quartile and slight overestimation at the 75\% quartile. Overall, TrajSurv’s calibration is competitive with other methods.

\subsection{Phenotypic Properties of Latent Space}
\begin{figure}[h]
    \centering
    \includegraphics[width=0.8\linewidth]{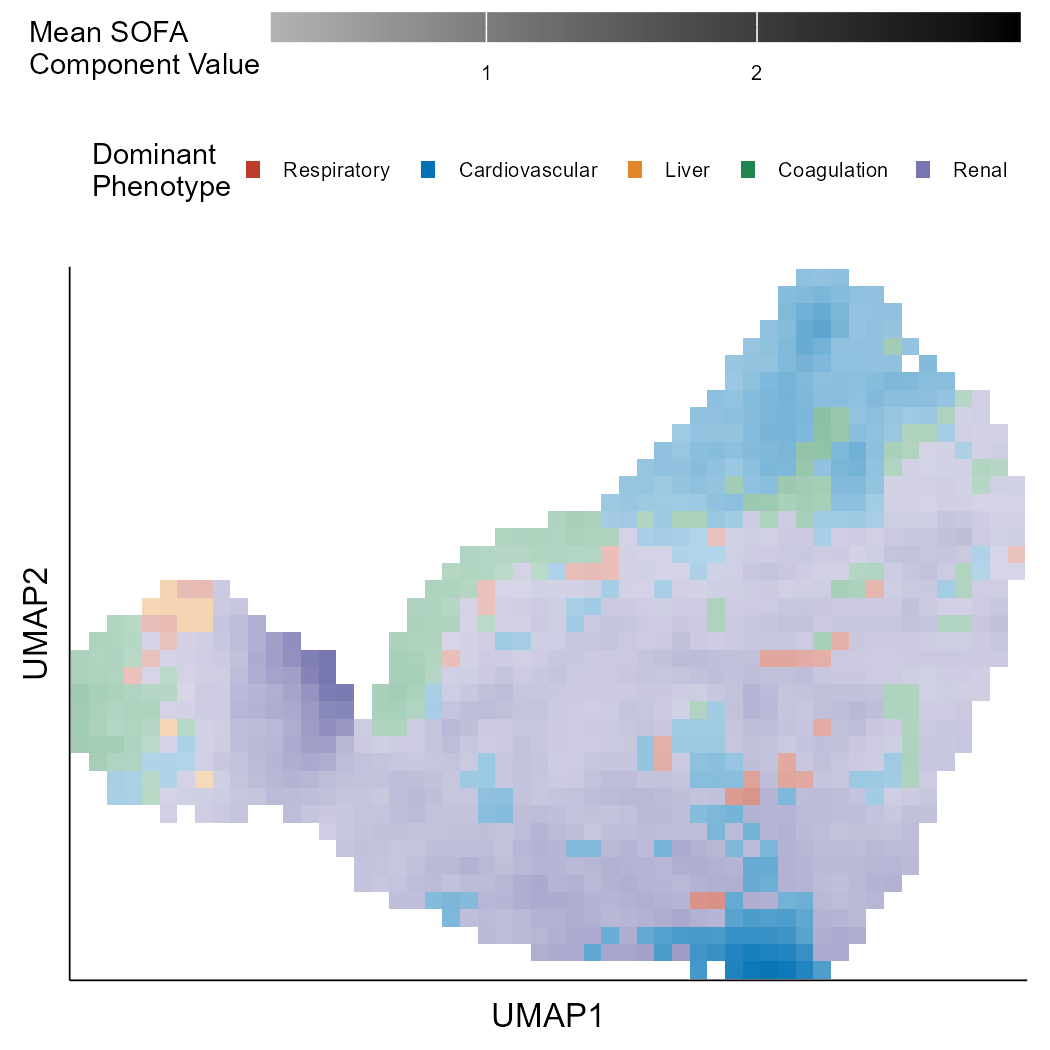}
    \caption{Dominant phenotypes in different regions of latent space. Different colors represent different dominant phenotypes, and the transparency of the color corresponds to the average severity (SOFA component) of the dominant phenotypes.}
    \label{fig:phenotype}
\end{figure}

To visually examine the phenotypic properties of the clinically aligned latent space, we plotted the dominant phenotypes and corresponding severity across regions in the latent space on MIMIC-III data. 
For each region, we averaged the SOFA component scores of the 50 nearest latent states, effectively representing the typical phenotype within that area. The dominant phenotype is then obtained from the highest average SOFA component in that region.
As shown in Figure \ref{fig:phenotype}, distinct regions in this space represent different clinical phenotypes, while close states reflect consistent dominant phenotypes and severity. For instance, the lower corner of the latent space corresponds to severe cardiovascular conditions. This demonstrates that the proximity in TrajSurv's latent space indicates similar patient phenotypic patterns. 
\subsection{Feature Importance and Relevance of All Features}
\begin{figure}[h]
    \centering
    \includegraphics[width=0.6\linewidth]{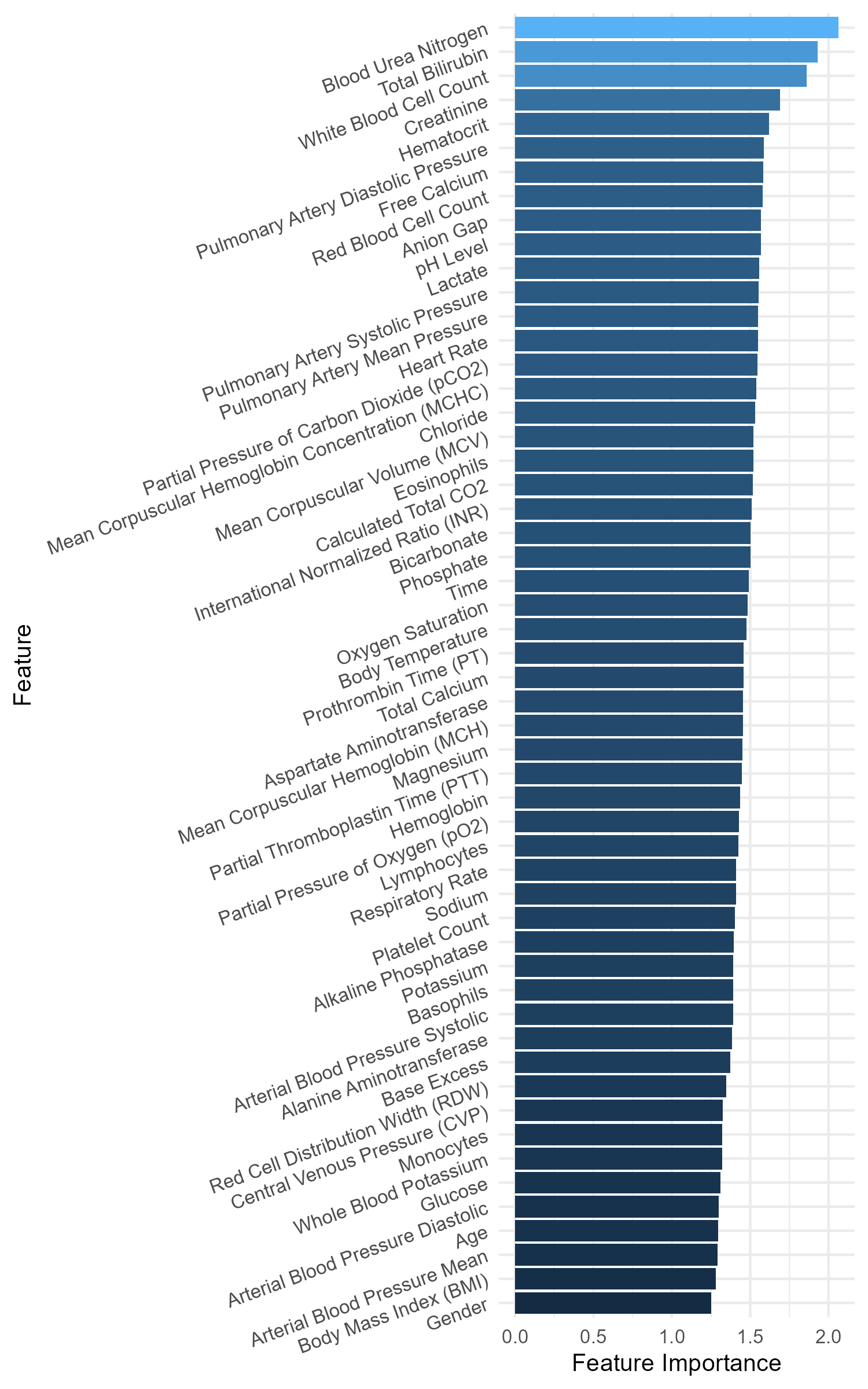}
    \caption{Feature importance of all clinical features.}
    \label{fig:full_importance}
\end{figure}

\begin{figure}[h]
    \centering
    \includegraphics[width=\linewidth]{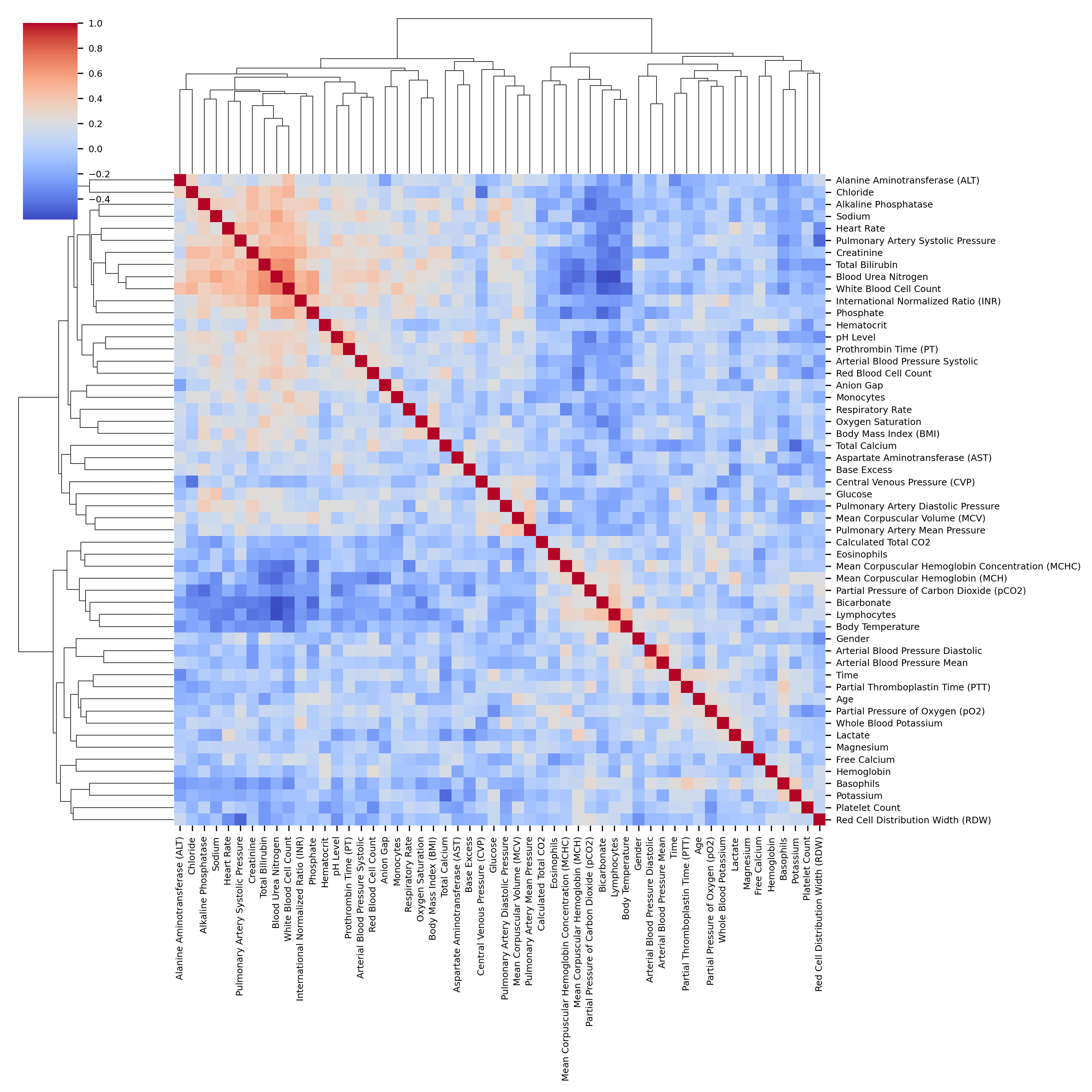}
    \caption{Heatmap of the cosine similarities between clinical features from the average vector field.}
    \label{fig:full_relevance}
\end{figure}

Figure \ref{fig:full_importance} and \ref{fig:full_relevance} show the full feature importance and relevance figures from the vector field interpretation on MIMIC III.
We note that these average patterns obtained from data-driven analysis should be carefully interpreted and should not be directly used in clinical practice.

\subsection{Interpretation Method Comparison}
TrajSurv's interpretation is a two-step process: (1) vector field interpretation for the feature-to-trajectory mapping, and (2) latent trajectory clustering for the trajectory-to-outcome linkage. The vector field interpretation offers insights distinct from, yet complementary to, methods like SHAP \cite{lundberg2017unified} or permutation importance \cite{altmann2010permutation}:
\begin{itemize}
    \item It explains how changes in clinical features dynamically influence the magnitude and direction of the patient's latent trajectory over continuous time.
    \item It allows assessment of feature relevance by measuring how similarly features drive this trajectory evolution via cosine similarity of vector field columns.
\end{itemize}

Table \ref{tab:interpretability_comparison} compares different interpretation methods. We highlight that TrajSurv combines vector field interpretation with latent trajectory clustering, hence enabling end-to-end interpretation from a dynamical perspective.

\subsection{Additional Experiments on Model Generalization}
\label{app:generalization}
We conducted additional experiments on TrajSurv's generalization using the MIMIC-III dataset. The results demonstrate TrajSurv's robustness across different settings.
\paragraph{Full vs. Reduced Data} We compared the full data (hourly data from the raw database) and the reduced data (only including hours with more than 20 observed features). The C-index is $0.806$ on the full data and $0.803$ on the reduced data, with about 3 times faster training on the reduced data. Therefore, we used reduced data during our experiments, which achieves computational efficiency while preserving fair comparison.
\paragraph{Short Trajectory} We performed an experiment using only the data between 18 and 36 hours after admissions, rather than the full 36-hour data in the main paper. TrajSurv achieves a C-index of $0.788$, compared to $0.803$ with 36-hour data, indicating robustness despite information loss.
\paragraph{Limited Training Data} To assess performance with limited training data, we experimented with a 20\% training, 40\% validation, and 40\% test split on MIMIC-III data. TrajSurv achieves a C-index of $0.752$, compared to $0.746$ of RSF, demonstrating generalizability even with reduced training data.
\subsection{Interpolation Method}
TrajSurv employs cubic Hermite splines with backward differences for input path interpolation. This is recommended by prior NCDE literature \cite{kidger2020neural,morrill2021neural} for its smoothness, online processing capability, and efficient integration. To explore alternatives, we experimented with rectilinear interpolation, which is also an online method, yielding a C-index of $0.789$, slightly below TrajSurv with cubic splines (C-index $0.803$).

% \begin{table}[h!] 
% \small
% \centering 
% \caption{Comparison of Interpretation Methods.} \label{tab:interpretability_comparison} \renewcommand{\arraystretch}{1.5} %
% \begin{tabularx}{\textwidth}{@{} l X X X @{}} 
% \toprule 
%  & \textbf{SHAP / Permutation Importance} & \textbf{Attention Mechanisms} & \textbf{TrajSurv's Vector Field Interpretation} \\ 
% \midrule
% \textbf{Primary Insight} & Contribution of each feature to a specific outcome & Which input features/time steps are most influential for an outcome & How feature changes drive the \textbf{evolution \& direction} of latent state trajectories \\ \addlinespace
% \textbf{Temporal Aspect} & Typically static for a given prediction; can be applied at different times but doesn't inherently model evolution & Highlights salient inputs/time steps for discrete predictions & \textbf{Models continuous-time dynamics}; captures how features influence on \textbf{trajectory evolution} \\ \addlinespace
% \textbf{Focus of Importance} & Importance for the magnitude of the final prediction & Importance for the final output & Importance for latent trajectory evolution (\textbf{magnitude and direction}) \\ \addlinespace
% \textbf{Feature Interaction} & Can compute SHAP interaction values & Does not directly provide feature interactions & \textbf{Relevance in co-directing trajectory evolution} \\  \addlinespace
% \textbf{Granularity} & Insight into outcome prediction & Insight into outcome prediction & Insight into the feature-to-trajectory \textbf{dynamical process} \\ 
% \bottomrule 
% \end{tabularx} 
% \end{table}

\begin{table}[h!] 
\small
\centering 
\caption{Comparison of Interpretation Methods.} 
\label{tab:interpretability_comparison} 
\renewcommand{\arraystretch}{1.5}
\begin{tabular*}{\textwidth}{@{\extracolsep{\fill}} l p{0.22\textwidth} p{0.22\textwidth} p{0.22\textwidth} @{}} 
\toprule 
 & \textbf{SHAP / Permutation Importance} & \textbf{Attention Mechanisms} & \textbf{TrajSurv's Vector Field Interpretation} \\ 
\midrule
\textbf{Primary Insight} & Contribution of each feature to a specific outcome & Which input features/time steps are most influential for an outcome & How feature changes drive the \textbf{evolution \& direction} of latent state trajectories \\ \addlinespace
\textbf{Temporal Aspect} & Typically static for a given prediction; can be applied at different times but doesn't inherently model evolution & Highlights salient inputs/time steps for discrete predictions & \textbf{Models continuous-time dynamics}; captures how features influence on \textbf{trajectory evolution} \\ \addlinespace
\textbf{Focus of Importance} & Importance for the magnitude of the final prediction & Importance for the final output & Importance for latent trajectory evolution (\textbf{magnitude and direction}) \\ \addlinespace
\textbf{Feature Interaction} & Can compute SHAP interaction values & Does not directly provide feature interactions & \textbf{Relevance in co-directing trajectory evolution} \\  \addlinespace
\textbf{Granularity} & Insight into outcome prediction & Insight into outcome prediction & Insight into the feature-to-trajectory \textbf{dynamical process} \\ 
\bottomrule 
\end{tabular*} 
\end{table}

\section*{Appendix B. Experimental Details}
\renewcommand{\thesubsection}{B.\arabic{subsection}}
\setcounter{subsection}{0}
\subsection{MIMIC-III Cohort}
\label{app:exp}
The MIMIC-III cohort includes 20,258 hospital admissions and 53 clinical features in 1-hour resolution. The dataset has 1,743 events. The 53 clinical features were the lab and chart events with top occurrence and demographics. The clinical features include arterial blood pressure diastolic, arterial blood pressure mean, arterial blood pressure systolic, alanine aminotransferase, alkaline phosphatase, anion gap, aspartate aminotransferase, base excess, basophils, bicarbonate, bilirubin total, calcium total, calculated total carbon dioxide, chloride, creatinine, central venous pressure, eosinophils, free calcium, glucose, heart rate, hematocrit, hemoglobin, international normalized ratio, lactate, lymphocytes, magnesium, mean corpuscular hemoglobin, mean corpuscular hemoglobin concentration, mean corpuscular volume, monocytes, oxygen saturation, pulmonary artery pressure diastolic, pulmonary artery pressure mean, pulmonary artery pressure systolic, partial pressure of carbon dioxide, pH, phosphate, platelet count, partial pressure of oxygen, potassium, potassium whole blood, prothrombin time, partial thromboplastin time, red cell distribution width, red blood cell count, respiratory rate, sodium, temperature, urea nitrogen, white blood cell count, gender, age, and body mass index.

The MIMIC-III cohort is inherently irregularly sampled. For computational feasibility and a fair comparison with baselines, we only included hourly time points where more than 20 features were observed. This cohort has an average of $4.02$ hourly clinical observations per patient within the first 36 hours and an overall feature missingness rate of 57.7\% across all time steps. 

\subsection{eICU Cohort}
The eICU cohort includes 116,503 hospital admissions and 53 clinical features in 1-hour resolution. The dataset has 7,958 events. The clinical features include basophils, eosinophils, lymphocytes, monocytes, polymorphonuclear leukocytes, alanine aminotransferase, aspartate aminotransferase, blood urea nitrogen, fraction of inspired oxygen, bicarbonate, hematocrit, hemoglobin, mean corpuscular hemoglobin, mean corpuscular hemoglobin concentration, mean corpuscular volume, mean platelet volume, prothrombin time, international normalized ratio, red blood cell count, red cell distribution width, white blood cell count (x1000), albumin, alkaline phosphatase, anion gap, bedside glucose, calcium, chloride, creatinine, glucose, magnesium, pH, partial pressure of carbon dioxide, partial pressure of oxygen, phosphate, platelet count (x1000), potassium, sodium, total bilirubin, total protein, temperature, arterial oxygen saturation, heart rate, respiration rate, central venous pressure, end-tidal carbon dioxide, systemic arterial systolic pressure, systemic arterial diastolic pressure, systemic arterial mean pressure, pulmonary artery systolic pressure, pulmonary artery diastolic pressure, pulmonary artery mean pressure, age, and body mass index.

The eICU cohort is also inherently irregularly sampled, and similarly preprocessed. This cohort has an average of $3.15$ hourly clinical observations per patient within the first 36 hours and an overall feature missingness rate of 55.0\% across all time steps.

\subsection{Hyperparameter Tuning}
Hyperparameters were tuned on the validation set. Specifically, for TrajSurv, we conducted a random search \cite{bergstra2012random} to tune the hyperparameters. The searching space of key hyperparameters is shown in Table \ref{tab:tune}. The search ranges were informed by common practice, prior work (e.g., 
$\kappa_1$ based on Rank-N-Contrast \cite{zha2024rank}), and empirical observations. Hyperparameters were tuned for the highest C-index on the validation set, with early stopping (patience 5) to prevent overfitting.

\subsection{Optimization}
We used AdamW \cite{loshchilov2019decoupledweightdecayregularization} to optimize TrajSurv. For the NCDE module, we used the \textit{torchcde} package with \textit{torchdiffeq} backend. We trained TrajSurv for 100 epochs, with early stopping based on the C-index on the validation set, with patience $5$. TrajSurv's training and evaluation were performed on a single NVIDIA Tesla T4 or RTX 2080 Ti GPU.

\subsection{Comparison Methods Implementation}
All baseline models underwent hyperparameter tuning using grid search on the validation set. For CoxPH, RSF, and Boosting, we used the implementations in the scikit-survival library \cite{sksurv}. For RDSM, we used the auton-survival package \cite{nagpal2022auton}. DDH and SLODE were implemented using the authors' original publicly available code.

\begin{table}[htbp]
\centering
\caption{Hyperparameter Tuning Ranges. \textbf{Bold} values indicate the final selected hyperparameters.}
\begin{tabular}{ll}
\toprule
Hyperparameter & Range \\
\midrule
$\alpha$ & $[0.75, \textbf{1.0}, 1.25]$ \\
$\kappa_1$ & $[1, \textbf{2}]$ \\
$\kappa_2$ & $[10, \textbf{30}, 50]$ \\
$\delta$ & $[10, \textbf{20}]$ \\
$d_z$ & $[32, \textbf{64}]$ \\
\bottomrule
\end{tabular}
\label{tab:tune}
\end{table}

\section*{Appendix C. More Background Information}
\renewcommand{\thesubsection}{C.\arabic{subsection}}
\setcounter{subsection}{0}

\subsection{Neural Controlled Differential Equations}
Neural controlled differential equations (NCDEs) extend neural ordinary differential equations (NODEs) to handle incoming data in an irregularly sampled time series setting. NCDEs leverage the mathematical framework of controlled differential equations (CDEs) to provide continuous-time modeling that dynamically adapts to data observations. A key advantage of NCDEs is their ability to utilize adjoint backpropagation for efficient training. The training of NCDEs has an overall memory footprint of $\mathcal{O}(L+H)$, where $L=t_n-t_0$ and $H$ is the memory footprint of the vector field. This contrasts previous work on NODEs for time series, which requires $\mathcal{O}(LH)$ memory. \cite{kidger2020neural}

\subsection{Dynamic Time Warping}
Dynamic Time Warping (DTW) is a technique used to measure the similarity between two time series of different lengths or varying speed. Previous work has also extended DTW to series of latent states, called dynamic state warping \cite{gong2017dynamicstatewarping}, which is similar to our latent trajectory clustering. The core of DTW involves constructing a cost matrix where each cell $(i, j)$ represents the distance between points in the two time series, and then finding the optimal warping path through this matrix. The accumulated cost matrix $D$ is built using the following recursive formula:

$$
D(i, j) = \text{cost}(i, j) + \min[D(i-1, j), D(i, j-1), D(i-1, j-1)],
$$
where $\text{cost}(i, j)$ is the distance between the $i$-th and $j$-th points of the two respective time series. The final DTW distance is the value of $D(n, m)$, where $n$ and $m$ are the lengths of the two time series.

In our latent trajectory clustering, the time series in DTW are the hourly extracted latent states from the continuous latent trajectories and the cost is the L2 distance between latent states.

\end{document}